\documentclass[letterpaper, 10 pt, conference]{ieeeconf}  

\IEEEoverridecommandlockouts                              

\usepackage{amsmath}
\usepackage{amssymb}
\usepackage{bm}
\usepackage{epsfig}
\usepackage{graphicx}
\usepackage{caption}
\usepackage{subcaption}
\usepackage{xcolor}
\let\labelindent\relax
\usepackage{enumitem}
\usepackage{acro}
\usepackage{algorithm}
\usepackage{tikz}
\usetikzlibrary{arrows.meta}
\usepackage{algpseudocode}
\usepackage[letterpaper=true,colorlinks,bookmarks=false]{hyperref}

\usepackage[backend=biber,style=ieee,doi=true,isbn=false,url=false,eprint=true,clearlang=true,date=year,citestyle=numeric-comp]{biblatex}
\AtEveryBibitem{%
    \clearname{editor}
    \clearfield{pages}
    \clearlist{publisher}
    \clearfield{series}
    \ifentrytype{inproceedings}{%
        \clearlist{address}
        \clearlist{location}
        \clearfield{volume}
    }{%
    }%
}

\title{\LARGE\bf Active Learning for UAV-based Semantic Mapping}

\author{Hermann Blum$^{1}$, Silvan Rohrbach$^{1}$, Marija Popovi\'{c}$^{1}$, Luca Bartolomei$^{2}$, Roland Siegwart$^{1}$\\\\$^{1}$Autonomous Systems Lab, ETH Z\"urich\\$^{2}$Vision for Robotics Lab, ETH Z\"urich
\thanks{This work was supported through funding by Hilti Group.}
}

\newcommand{\Par}[1]{\vskip4pt \noindent{\bf #1~}}

\usepackage{xspace}
\DeclareRobustCommand\onedot{\futurelet\@let@token\@onedot}
\def\onedot{\ifx\@let@token.\else.\null\fi\xspace}

\def\eg{\emph{e.g}\onedot} 
\def\ie{\emph{i.e}\onedot}

\def\etal{\emph{et al}\onedot}

\DeclareAcronym{uav}{
  short = UAV,
  long  = unmanned aerial vehicle,
}
\DeclareAcronym{ood}{
  short = OoD,
  long  = out-of-distribution,
}
\DeclareAcronym{ipp}{
  short = IPP,
  long  = informative path planning,
}
\DeclareAcronym{cnn}{
  short = CNN,
  long  = convolutional neural network,
}
\DeclareAcronym{mIoU}{
  short = mIoU,
  long  = mean intersection over union,
}

\begin{document}

\maketitle
\thispagestyle{empty}
\pagestyle{empty}

\begin{abstract}
\Aclp*{uav} combined with computer vision systems, such as convolutional neural networks, offer a flexible and affordable solution for terrain monitoring, mapping, and detection tasks. However, a key challenge remains the collection and annotation of training data for the given sensors, application,  and mission. We introduce an \acl*{ipp} system that incorporates novelty estimation into its objective function, based on research for uncertainty estimation in deep learning. The system is designed for data collection to reduce both the number of flights and of annotated images. We evaluate the approach on real world terrain mapping data and show significantly smaller collected training dataset compared to standard lawnmower data collection techniques.
\end{abstract}

\section{Introduction}
\Acp{uav} offer cost-efficient, flexible and automated delivery of high-quality sensing data in various applications, \eg\ search-and-rescue, inspection and agricultural monitoring. Image sensors in particular are well suited for \ac{uav} based sensing because of their low cost, size, and power demand. Recent advances in computer vision and deep learning enable automated analysis of the gathered data and make the system applicable to a range of monitoring and mapping tasks in large or hard-to-access environments.

In this work, we look into the problem of data collection for environmental monitoring. Especially in mentioned applications like long-term agricultural monitoring, imagery can change drastically over time and differ from the training data. Cases of novelty require tedious data collection, annotation and retraining, with two time-costly challenges: On the one hand the flight time is limited due to the energy consumption of the \ac{uav}, on the other hand the time investment for manually annotating the large pool of acquired images is huge. By flying over an area in a conventional predetermined lawnmower-fashion, energy is wasted on gathering repetitive and similar images that will not significantly improve the segmentation and classification quality. We propose an \ac{ipp} system that actively searches for and gathers data different to the training distribution of the available semantic segmenter.

Based on results from the recent `Fishyscapes' benchmark for novelty detection in semantic segmentation~\cite{Blum2019-eh}, we propose an \ac{ipp} system that maximises the novelty of the gathered images in a single flight mission. Given the available resources, we reduce the number of flights and annotated data, while achieving faster improvements of the semantic segmentation.

\begin{figure}[t]
    \centering
    \begin{tikzpicture}
\tikzstyle{vecArrow} = [thick, decoration={markings,mark=at position
   1 with {\arrow[semithick, fill=black]{open triangle 60}}},
   double distance=1.4pt, shorten >= 5.5pt,
   preaction = {decorate},
   postaction = {draw,line width=1.4pt, white,shorten >= 4.5pt}]
\tikzstyle{myarrows}=[line width=1mm,draw=blue,-triangle 30,postaction={draw, line width=1mm, shorten >=4mm, -}]
   
\node[inner sep=0pt] at (0,0) {\includegraphics[width=90pt]{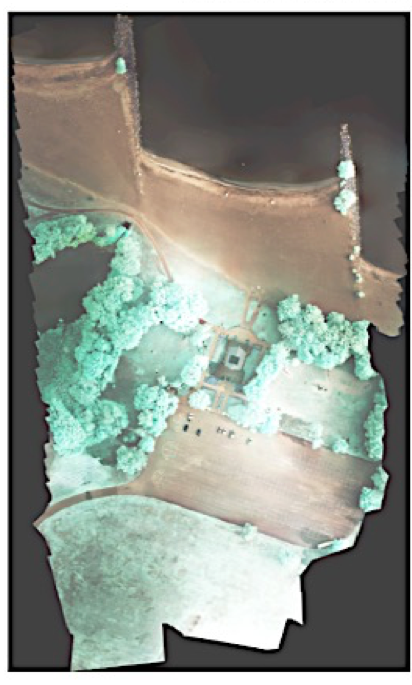}};
\node[inner sep=0pt] at (0.2,0.1) {\includegraphics[width=85pt]{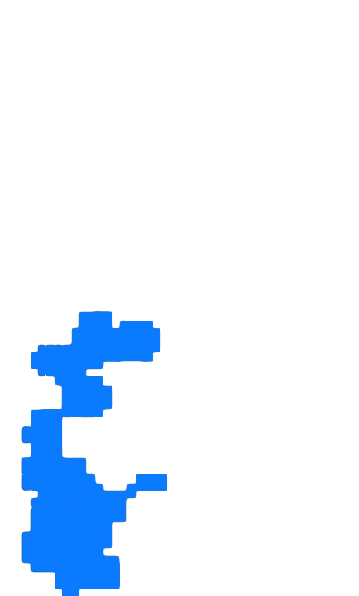}};

\node[inner sep=0pt] (drone) at (0,-0.1) {\includegraphics[width=20pt]{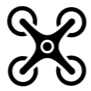}};
\node[right of=drone] (a) {};
\node[above of=drone] (b) {};
\node[left of=drone] (c) {};
\node[below of=drone] (d) {};
\draw[line width=2pt, -{Latex[length=3mm]}] (drone)--(a);
\draw[line width=2pt, -{Latex[length=3mm]}] (drone)--(b);
\draw[line width=2pt, -{Latex[length=3mm]}] (drone)--(c);
\draw[line width=2pt, -{Latex[length=3mm]}] (drone)--(d);

\node[inner sep=0pt, label=drone image] (image) at (3,0) {\includegraphics[width=50pt]{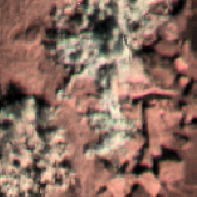}};
\node[inner sep=0pt, label=novelty] (heatmap) at (5,0) {\includegraphics[width=50pt]{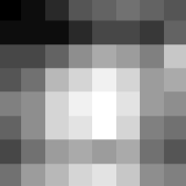}};

\draw (drone.center) -- (image.north west);
\draw (drone.center) -- (image.south west);
\draw (drone.center) -- (image.north east);
\draw (drone.center) -- (image.south east);

\node[inner sep=0pt] at (image) {\includegraphics[width=50pt]{plots/droneimage.png}};
    \end{tikzpicture}
    \caption{To collect useful training data for a semantic segmentation network, our \acl{ipp} algorithm finds a path for the drone that increases the novelty of the observed images. The novelty is evaluated in a patchwise manner that additionally gives directional information from the gradient over the heatmap. Brighter novelty is higher. The planned path is marked in blue.}
    \label{fig:teaser}
\end{figure}

An illustration of our approach is given in figure~\ref{fig:teaser}. For every new image captured by the \ac{uav}, we estimate the novelty of the image and follow this information to high-interest regions.

The contributions of this work are the following:
\begin{itemize}
    \item We propose an \ac{ipp} algorithm that uses novelty estimation from deep learning as primary source of information.
    \item We evaluate the proposed \ac{ipp} solution and the novelty estimation towards the problem of active learning.
\end{itemize}

\begin{figure*}[ht]
   \centering
   \includegraphics[width=\linewidth]{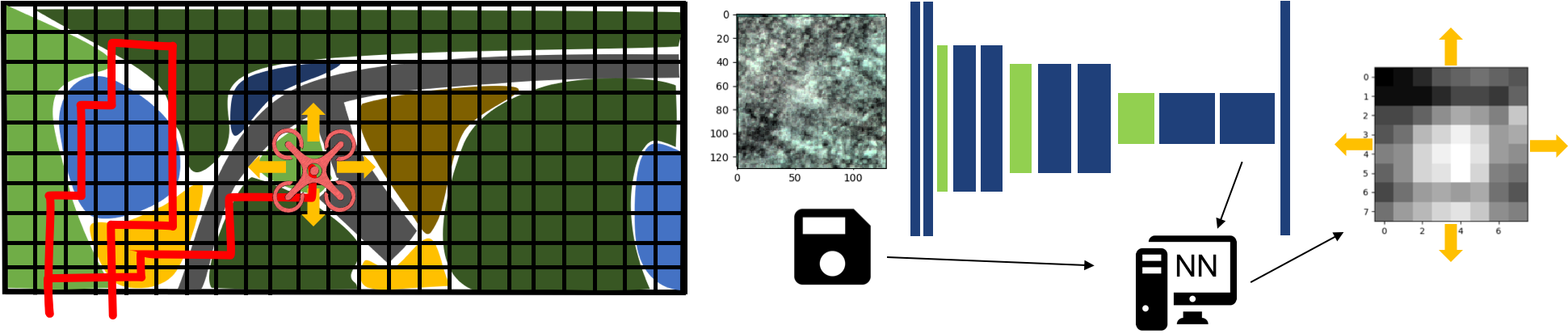}
   \caption{Illustration of the path planning scenario. \textbf{Left.} The \ac{uav} is in between its mission to collect useful new training data. The path is planned as a sequence of grid cells on the map. \textbf{Right.} The current image seen from the \ac{uav} is put through the \ac{cnn} to get a heatmap of patchwise novelty scores. The heatmap is then used to select the most promising next gridcell.}
   \label{fig:DiscGrid}
\end{figure*}

\section{Related Work}
\label{sec:related_work}

\Par{Novelty detection} or uncertainty estimation for deep learning models is a very active area of research. Uncertainty and mistakes in prediction algorithms can come from noise in the input, but also from differences between the training data and the input. In our case, we are exclusively interested in the second part, as we want to decrease the distance between input and training distributions by gathering a broader range of training data with our \ac{ipp} setup.\\
Hendrycks\ \etal~\cite{Hendrycks2017-ua} give a comprehensive overview of the problem and evaluate the baseline of the softmax output. Bayesian Deep Learning~\cite{Gal2016-mx,Kendall2017-jy} estimates uncertainty from deep models whose outputs and weights are probability distributions. Different works compare the flow of data through the network to the training data and estimate uncertainty as deviation from the training distribution~\cite{Mandelbaum2017-ti,Papernot2018-xz}. \cite{Blum2019-eh} adapted and evaluated these approaches towards novelty detection in semantic segmentation. As a third direction of research, learned representations of the input are reconstructed and compared in input space~\cite{Pidhorskyi2018-qf,Lis2019-vd}.

\Par{Active Sampling} is known in machine learning as a technique for data reduction to speed up training times. Wang\ \etal~\cite{Wang2017-tz} show a system that samples from training images based on the softmax confidence. Gal\ \etal~\cite{Gal2017-hn} develop a similar system based on Bayesian Deep Learning, which is more suited for novelty detection. Both systems focus on reducing the expensive labelling and do not take into account the problem of data acquisition.

\Par{Informative Path Planning} has recently experienced increasing interest for a variety of applications, such as environmental monitoring~\cite{Hitz2017-kh, Hollinger2014-md, Ghaffari2019}, surveillance~\cite{vivaldini2016} and inspection~\cite{Bircher2018}. The aim is efficient continuous or discrete data acquisition in complex environments using a mobile robot, whose motions are constrained by its sensing and mobility capabilities. Popular techniques include Partially Observable Markov Decision Processes (POMDPs)~\cite{Kaelbling1998} or Gaussian Processes~\cite{Rasmussen2005}. However, the \ac{uav} sensing scenario represents an extreme case where only information about past places in the trajectory is available, which makes path planning for more than a few steps unfeasible and reduces the applicability of the mentioned methods.\\
While they do not use novelty as input to a path planner, Richter~\etal~\cite{Richter2017-wg} proposed one of the first path planning systems with novelty detection based on deep learning. They use reconstruction based novelty to safely switch between a neural network-based obstacle avoidance controller and a slower conservative planner. In our method, we directly exploit the novelty information to steer the robot towards more informative regions.

\section{Method}
\label{sec:method}

\subsection{Novelty Detection}
Our novelty estimation approach follows results from~\cite{Mandelbaum2017-ti,Papernot2018-xz,Blum2019-eh} and uses density estimation in the feature space. In particular, we use  kernel density estimation to produce patchwise uncertainty estimates. Lower resolution feature vectors from a convolutional neural network are compared to their nearest neighbors from the training distribution. Novelty is here defined as the average cosine distance to neighbors from the training distribution, a metric that was shown to work well in different scenarios in the aforementioned works. However, we note that the proposed system setup is independent of the underlying technique for novelty detection as long as it produces pixel- or patchwise values.

Given a training set of images $\bm{A}$, we extract embeddings $\bm{Z}_l = f_l(\bm{A})$ at layer $l$ from the segmentation network and store them in a database. For a given input image $\bm{a}'$ and embeddings $\bm{z}_l' = f_l(\bm{a}')$, we then approximate the probability density of the input image with respect to the distribution of training data using a kernel density estimation of $\bm{z}_l'$ with respect to the $k$ nearest neighbors in $\bm{Z}_l$. This can be found through
\begin{align*}
    D(\bm{z}') = \sum \limits_{i=0}^{k-1} \frac{\bm{z}'\bm{z}^{(i)}}{|\bm{z}'|\,|\bm{z}^{(i)}|}.
\end{align*}

$D(\bm{z}')$ is a patchwise uncertainty estimation of the current input image. The size of the patches is dependent on the layer $l$ where the embeddings are extracted, \ie\ usually the resolution is 8 or 16 times smaller than the input image. An example of the uncertainty estimation is shown in figure~\ref{fig:teaser}.

For our input image size, the above approach requires 64 nearest neighbor searches per input image and is therefore not feasible for real-time. However, it can be directly switched out with flow-based density estimation as was very recently shown in~\cite{Blum2019-eh}, which only requires a single pass through a network.

\subsection{Path Planning}
The objective in our \ac{ipp} problem is to find new, informative images for training. The difficulty of approaching this problem is twofold. First, the information computed by the novelty detection is available only in locations that have already been visited by the \ac{uav}; second, the distribution of novelty over the explorable space is unknown. In this work, we adopt the assumption of spatial correlation, \ie\ while we do not pose any hypothesis on where to find novel inputs, we assume that there are regions of connected novelty cells scattered over the map, rather than an i.i.d. uniform distribution.\\
Given a grid discretization of the world, the path-planning problem is stated as active sampling from the adjacent cells of the current position in the grid map. An illustration of the problem is given in figure~\ref{fig:DiscGrid}.

At each re-planning step, the following information is available to the path planner:
\begin{itemize}
    \item number of additional explorable cells (\ie\ battery life);
    \item explored cells in the map;
    \item distances to the borders of the explorable world;
    \item average novelty score of the current cell image;
    \item gradient direction of the novelty score from the current cell image.
\end{itemize}

We implement the path planning on basis of potential fields, where each grid cell $\bm{p}$ on the map has an assigned potential $\varphi(\bm{p})$:
\begin{align*}
    \varphi(\bm{p}) &= D(\bm{p}) + \textrm{penalty}_\textrm{border}(\bm{p}) + \textrm{penalty}_\textrm{visited}(\bm{p})
\end{align*}

The novelty of a grid cell image $D(\bm{p})$ is initialized uniformly to a constant value for all unknown patches and updated once a patch has been observed. $\textrm{penalty}_\textrm{border}(\bm{p})$ increases the potential towards the border of the observable world. This term is required only in simulated environments, in order to avoid situations where the \ac{uav} gets stuck in corners of the simulation environment. $\textrm{penalty}_\textrm{visited}(\bm{p})$ is a constant penalty applied to cells that have already been visited and therefore should be avoided.

\begin{figure}[t]
    \centering
    \sffamily
    \begin{tikzpicture}
\node[inner sep=0pt] (heatmap) at (0,0) {\includegraphics[width=40pt]{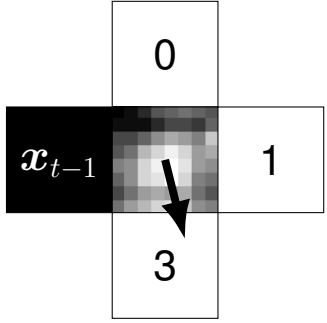}};
\node[draw, inner sep=0pt, minimum size=40pt] (right) at (40pt,0) {\LARGE 1};
\node[draw, inner sep=0pt, minimum size=40pt] (top) at (0,40pt) {\LARGE 0};
\node[draw, inner sep=0pt, minimum size=40pt, fill=black] (left) at (-40pt,0) {\color{white}\LARGE \strut$\bm{x}_{t-1}$};
\node[draw, inner sep=0pt, minimum size=40pt] (bottom) at (0,-40pt) {\LARGE 3};

\node[] (gradient) at (0.3,-1.2) {};
\draw[line width=3pt, -{Latex[length=15pt]}] (heatmap.center)--(gradient);

    \end{tikzpicture}
    \caption{Example of the gradient propagation. The \ac{uav} came from cell $\bm{x}_{t-1}$ and obtained an image in the center cell $\bm{x}_t$, which is used to produce the shown novelty heatmap. As the novelty of the adjacent cells is still unknown, we infer information from the gradient (black arrow) of the novelty heatmap $\nabla D(\bm{x}_t)$. A score is assigned to each of the neighboring cells and is then substracted from the potential field.}
    \label{fig:gradient_propagation}
\end{figure}

\begin{algorithm}[ht]
\caption{Selection of the next adjacent grid cell}\label{alg:IPP}
\begin{algorithmic}[1]
\State $t \gets 0$
\While{enough energy to head home}
	\If{surrounded by visited patches}
		\State follow shortest path to non-visited patch
	\EndIf
	\Statex
	\State update $\varphi(\bm{x}_t)$ with observed novelty
	\State $\hat{\varphi}(\bm{p}) \gets \varphi(\bm{p})$ \Comment{temporary updates to $\varphi$}
    \Statex
	\If {$D(\bm{x}_t) < \alpha$} \Comment{low novelty}
	    \Statex \Comment{no updates to $\hat{\varphi}$}
	\EndIf
    \Statex
	\If {$\alpha < D(\bm{x}_t) \leq \beta$} \Comment{medium novelty}
	    \State $\hat{\varphi}(\bm{p}) \gets \hat{\varphi}(\bm{p}) - f(\bm{p}, \nabla D(\bm{x}_t))$
	    \Statex \Comment{propagate novelty gradient, see fig.~\ref{fig:gradient_propagation}}
	\EndIf
	\Statex
	\If {$\beta < D(\bm{x}_t)$} \Comment{high novelty}
	    \State $\hat{\varphi}(\bm{p}) \gets \hat{\varphi}(\bm{p}) \circledast {\tiny\frac{1}{9}\begin{bmatrix}1\hspace{-7pt} & 1\hspace{-7pt}& 1 \\ 1\hspace{-7pt} & 1\hspace{-7pt}& 1 \\ 1\hspace{-7pt} & 1\hspace{-7pt} & 1\end{bmatrix}}$
	    \Comment{smooth potential}
	\EndIf
    \Statex
    \If {$D(\bm{x}_{t-1}) > \beta$ and $D(\bm{x}_t) < \beta$}
    \Statex\Comment{edge of informative region}
    \State $\hat{\varphi}(\bm{x}_\textrm{straight}) \gets \hat{\varphi}(\bm{x}_\textrm{straight}) + \textrm{penalty}_\textrm{forward}$
    \Statex\Comment{don't go forward, away from high novelty}
    \EndIf
	\Statex
	\State choose direction with lowest $\hat{\varphi}(\bm{p})$
	\State $t \gets t + 1$
\EndWhile
\State return home
\end{algorithmic}
\end{algorithm}

Based on the defined potential field, the drone selects one of the 4 adjacent grid cells to its current location $\bm{x}_t$ according to algorithm~\ref{alg:IPP}. The algorithm follows a scheme of fast traversion of low-novelty areas and exhaustive exploration of high-novelty areas. We distinguish 3 different cases, depending on the novelty of position $\bm{x}_t$:
\begin{description}
    \item[low novelty] is defined as $D(\bm{x}_t) \leq \alpha$. These are images uninteresting for training. The path planner tries to avoid these regions and follows the gradient of the potential field towards more informative areas.
    \item[medium novelty] is defined by $\alpha < D(\bm{x}_t) \leq \beta$. It can depict borders between different classes, as well as borders to more informative areas. The path planner takes the gradient of the current novelty heatmap as additional information into account, as it might be directed towards high novelty regions.
    \item[high novelty] is defined by $D(\bm{x}_t) > \beta$. It identifies regions that contain crucial data for training. Instead of following the gradient, we found that it is more helpful to explore larger areas of high-average novelty. In order to encourage exploration, we smooth the potential field to not disturb the path planner with local fluctuation of the novelty estimation function. Moreover, we add $\textrm{penalty}_\textrm{forward}$ for moving out of high-novelty regions.
\end{description}
We set $\alpha$ and $\beta$ based on the lower and upper quartile thresholds on a validation set. The overall goal of the \ac{ipp} algorithm is to catch as many high-novelty cells in a mission as possible.

\section{Evaluation}
\label{sec:evaluation}
We evaluate our approach on the remote sensing task of the RIT18~\cite{Kemker2018-kc} dataset. The dataset contains  high-resolution hyperspectral images of the same location for two points in time, suitable as a training and validation dataset. For both images, ground truth annotations are given. We use the classes \emph{grass}, \emph{tree}, \emph{beach}, and \emph{other}. We simulate the \ac{uav} flight by laying a grid over the image with cell size $128 \times 128 \textrm{ px}$.

The images gathered from each grid cell are segmented using a fully convolutional network~\cite{Long2015-fm} with a VGG-16 encoder~\cite{Simonyan2015-zf}, in particular the implementation from~\cite{Blum2018-kp}. We use the embeddings from the \emph{conv5-1} layer and sample the density over 20 nearest neighbors.

The experiment is set up as follows. The \ac{uav} is sent on multiple missions, each time with the objective to gather new training data. After each mission, we add the new images together with annotations to the pool of training data, retrain the semantic segmenter, and build a new kNN database. The new semantic segmenter is then used for the novelty estimation in the next mission.\\
At every iteration we evaluate the accuracy of the segmenter on the whole map measured in \ac{mIoU}.

To evaluate our approach, we compare against two different lawnmower baselines:
\begin{description}
\item[a - big lawnmower] For the conventional approach flights with the UAV are simulated in lawnmower fashion across the whole site. A starting point near the edge of the site is chosen and from there line after line, back and forth across the map images are acquired until there is no energy left and the (re-)training of the network is performed.
\item[b - small lawnmower] For more diverse image acquisition with less flights, the small-scale lawnmower approach works by manually choosing different starting positions spread out across the whole site. From each of the starting positions a flight with a small-scale lawnmower approach is executed which results in the collection of imagery in a rectangle with a size depending on the energy capacity of the UAV. After each flight, the gathered image patches are then used for (re-)training. 
\item[c - \acl{ipp}] To test the developed IPP system the same starting positions as the one used for the small lawnmower approach are used. Instead of predetermined paths, from the second flight on the IPP system is used to guide the UAV autonomously until the energy is depleted, then the vehicle heads back to the starting position and the network is (re-)trained with the acquired images. 
\end{description}

\begin{figure}
    \centering
    \includegraphics[width=\linewidth]{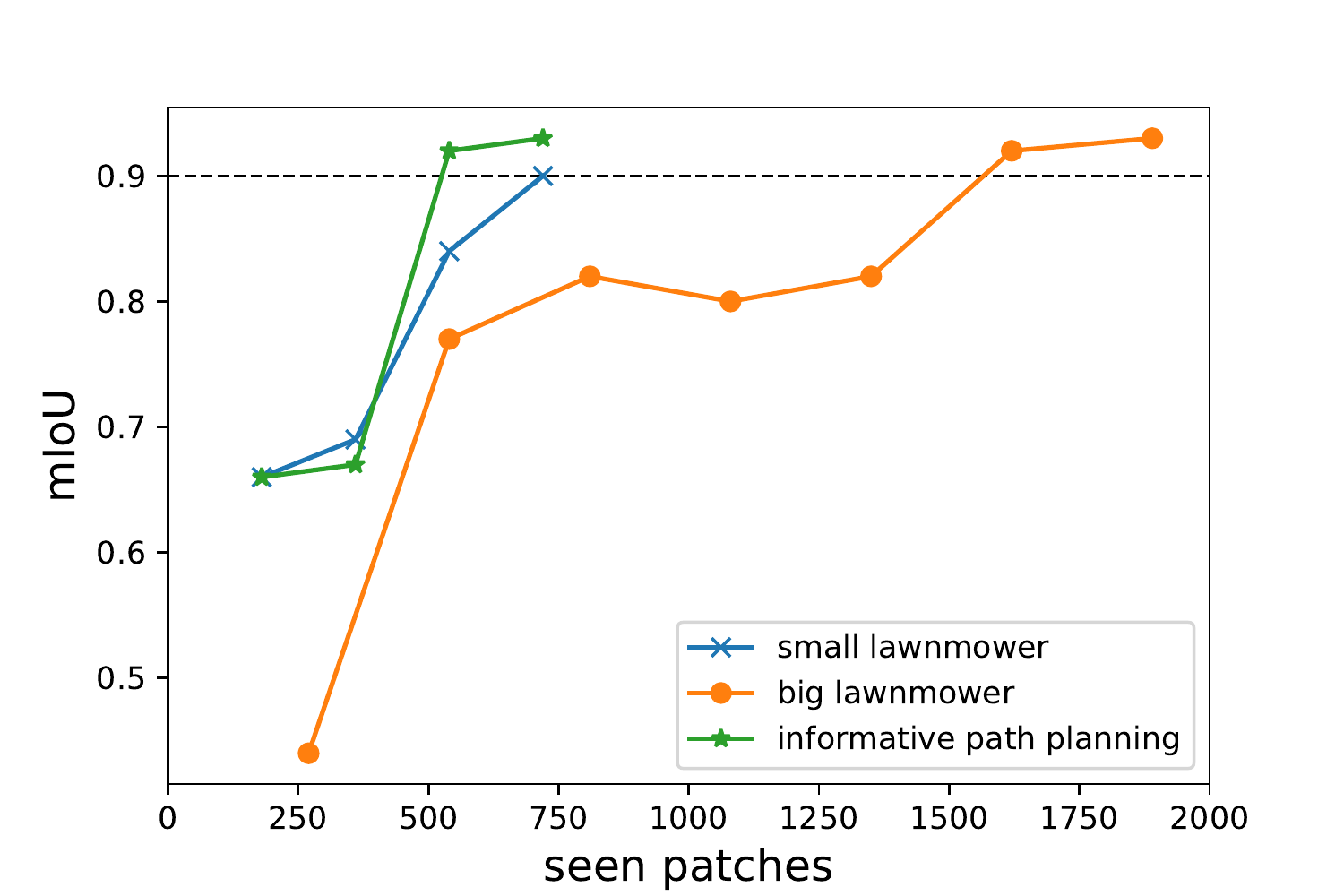}
    \caption{Comparison of the mean IoU on the full map after every retraining of the networks following the three different approaches.}
    \label{fig:iou_plot}
\end{figure}

\begin{figure}[t!]
    \centering
    \footnotesize
    \def\iwidth{.28\linewidth}
    \begin{subfigure}[b]{\linewidth}
\hspace{2pt}\begin{tikzpicture}
    \node[inner sep=0pt, label=path] at (0,0) {\includegraphics[width=\iwidth]{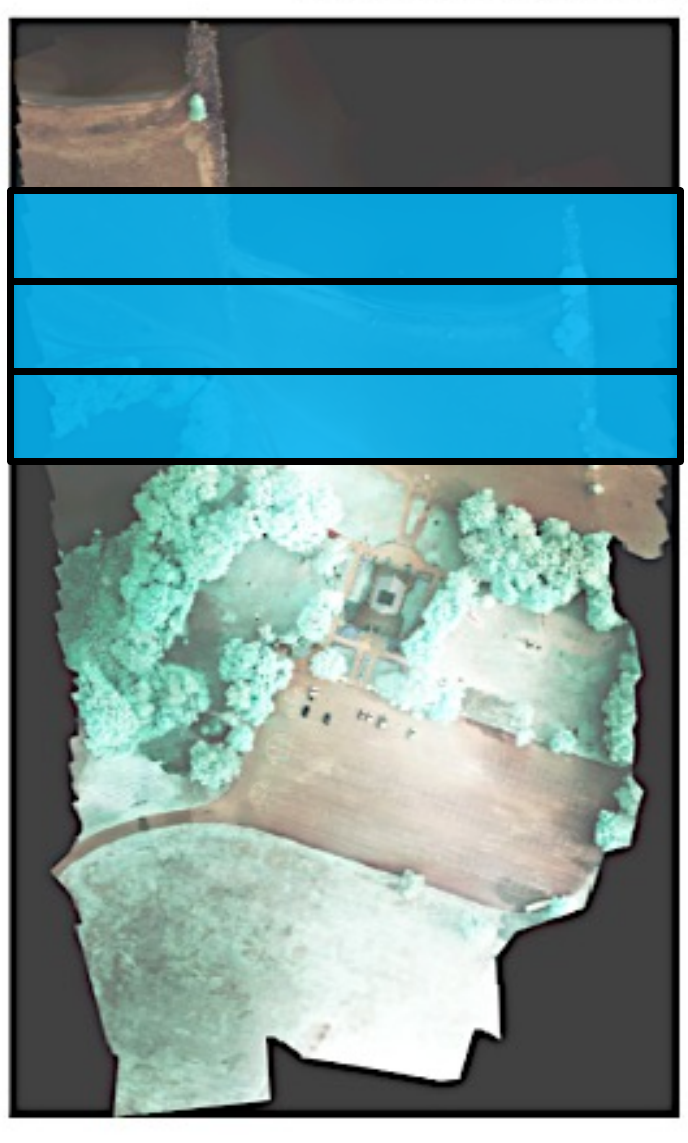}};
    \node[inner sep=0pt, label=prediction] at (3,0) {\includegraphics[width=\iwidth]{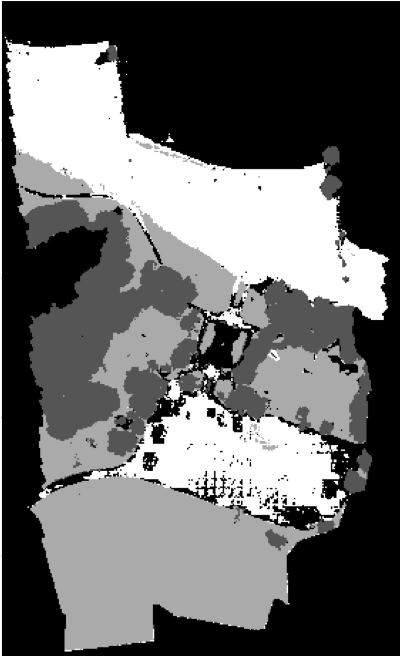}};
    \node[inner sep=0pt, label=ground truth] at (6,0) {\includegraphics[width=\iwidth]{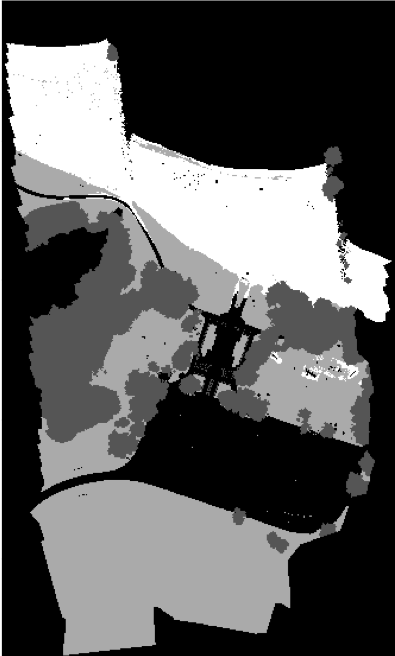}};
\end{tikzpicture}
         \caption{The big lawnmower is started in the upper half of the map and goes down row by row over the map.}
         \label{fig:planned_paths:a}
     \end{subfigure}
     \hfill
     \begin{subfigure}[b]{\linewidth}
     \hspace{2pt}
\begin{tikzpicture}
    \node[inner sep=0pt, label=path] at (0,0) {\includegraphics[width=\iwidth]{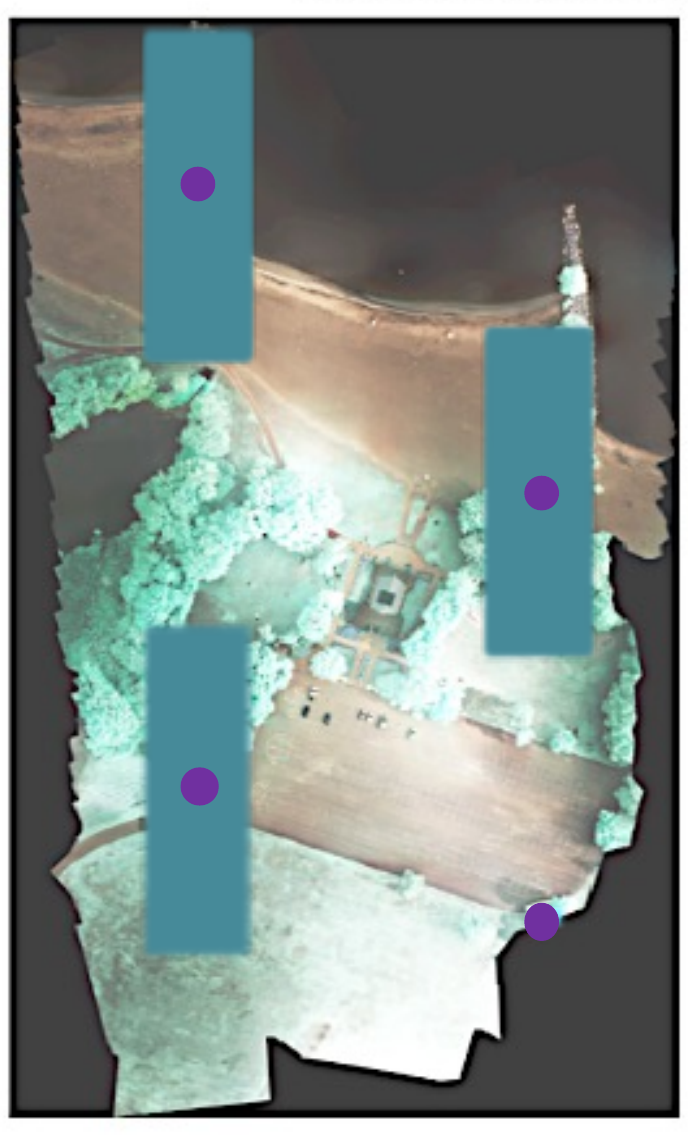}};
    \node[inner sep=0pt, label=prediction] at (3,0) {\includegraphics[width=\iwidth]{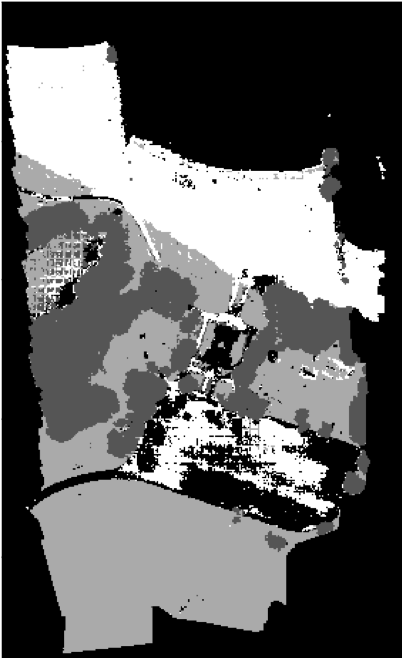}};
    \node at (5.4, 0) {\begin{tikzpicture}
        \node[inner sep=0pt, label=right:sand] at (0,-0.7) {\includegraphics[width=8pt]{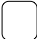}};
        \node[inner sep=0pt, label=right:tree] at (0,-1.1) {\includegraphics[width=8pt]{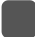}};
        \node[inner sep=0pt, label=right:grass] at (0,-1.5) {\includegraphics[width=8pt]{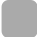}};
        \node[inner sep=0pt, label=right:other] at (0,-1.9) {\includegraphics[width=8pt]{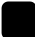}};
    \end{tikzpicture}};
\end{tikzpicture}
         \caption{As a better comparison to the \ac{ipp} experiment, we start small lawnmower missions at different locations distributed over the whole map. Each mission starts at one of the blue dots and executes a rectangular pattern around this position.}
         \label{fig:planned_paths:b}
     \end{subfigure}
     \hfill
     \begin{subfigure}[b]{\linewidth}
     \hspace{2pt}
\begin{tikzpicture}
    \node[inner sep=0pt, label=path] at (0,0) {\includegraphics[width=\iwidth]{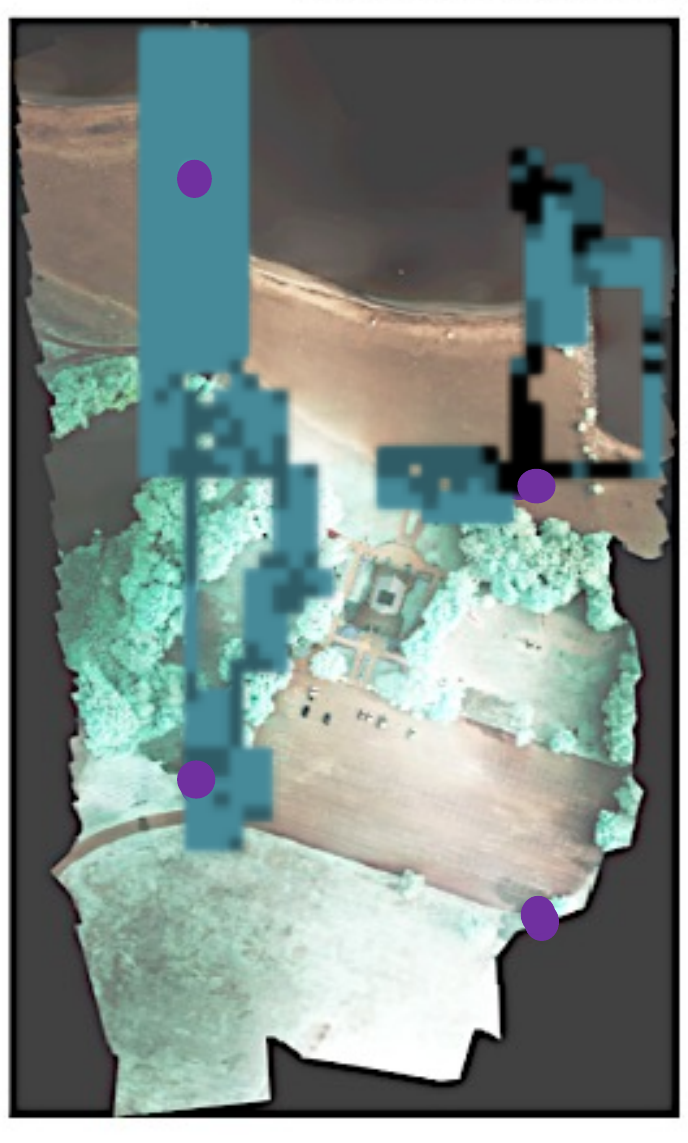}};
    \node[inner sep=0pt, label=prediction] at (3,0) {\includegraphics[width=\iwidth]{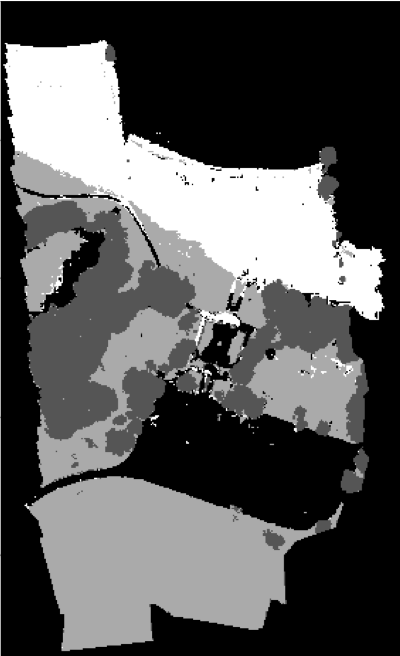}};
    \node[inner sep=0pt, label=novelty] at (6,0) {\includegraphics[width=\iwidth]{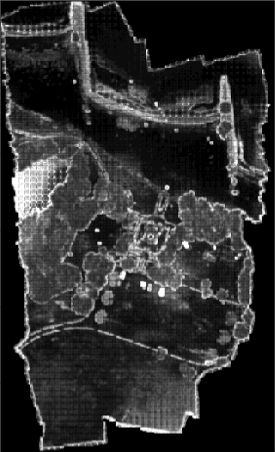}};
\end{tikzpicture}
\caption{The first three missions with \ac{ipp}. Note that the first mission has no prior data available, cannot estimate novelty and therefore falls back to the small lawnmower approach. For the second and third path, low novelty cells are marked dark and medium novelty cells are marked in light gray. On the right we show the novelty corresponding to the prediction after the first three missions (higher novelty is brighter). This information is not available to the planner as novelty can only be evaluated for visited places.}
         \label{fig:planned_paths:c}
     \end{subfigure}
     \caption{For each path planner, we show the observed grid cells after three missions in blue (left) and the semantic segmentation over the whole map, trained on the gathered training data from the first three missions (center).}
     \label{fig:planned_paths}
\end{figure}

The experimental results are shown in figure~\ref{fig:iou_plot}. The experiment validates that our \ac{ipp} system collects useful data faster than the lawnmower based approaches. To reach a good segmentation performance of $\textrm{mIoU}\geq 90\,\%$, only 3 \ac{uav} missions were necessary. As a comparison, the prediction maps and generated paths after 3 missions of all methods are shown in figure~\ref{fig:planned_paths}. In particular, we note that the novelty plotted over the whole map after 3 missions in~\ref{fig:planned_paths:c} highlights the lake on the upper left as a region of high novelty, which is also the only region that is wrongly classified. The corresponding path shows that the \ac{ipp} was aborted before collecting more lake data due to energy constraints and went back to the starting position.

\section{Discussion}
In our experiments with the RIT18 dataset, we found that one of the main differences among the evaluated path planners was the data balance of the different classes captured in each mission, because the classes are geographically very separated on the map. This makes the experiment in general sensitive to the choice of starting positions for each mission. Choosing positions in all 4 quadrants of the map came to the advantage of the smaller lawnmower approach for the particular dataset. It remains to test how our \ac{ipp} framework performs with a starting position fixed over several missions. To disentangle the evaluation of our \ac{ipp} system and the novelty detection, we plan experiments on datasets with a different class distribution, as well as experiments where we exchange the novelty estimation to a randomly generated heatmap.

On top of the points above, we restricted the path planning problem by the available information and the possible actions. For other scenarios, the information available can vary, \eg\ it is also possible to have novelty maps similar to the one on figure~\ref{fig:planned_paths:c} from previous missions available. In our experiments the height above ground was kept constant. While informative path planning with variable height was explored in different works before, it remains subject of further research how the height affects segmentation performance and novelty estimation.

\section{Conclusion}
In this work we present an \ac{ipp} system to collect valuable training data with a \ac{uav}. We show how to incorporate novelty estimation from deep learning into a path planning objective and evaluate our system on a real world terrain monitoring map. The results indicate a significantly faster useful data acquisition with improved performance compared to traditional lawnmower approaches.
\section*{Acknowledgment}
We thank Cesar Cadena and Juan Nieto for their valuable inputs.

\printbibliography

\end{document}